\pdfoutput=1

\documentclass[11pt]{article}

\usepackage[]{ACL2023}

\usepackage{makecell}
\usepackage{xspace}
\usepackage{hyperref}
\usepackage{arydshln}
\usepackage{times}
\usepackage{latexsym}
\usepackage{amssymb}
\usepackage[T1]{fontenc}

\usepackage[utf8]{inputenc}
\usepackage{multirow}

\usepackage{microtype}

\usepackage{inconsolata}
\usepackage{graphicx}
\usepackage{booktabs}
\usepackage{subcaption}
\usepackage{xcolor}
\usepackage{enumitem}
\usepackage{bm}
\newcommand{\name}{PEMT\xspace}

\title{\name: Multi-Task Correlation Guided Mixture-of-Experts Enables Parameter-Efficient Transfer Learning}

\author{
  Zhisheng Lin\textsuperscript{1},  Han Fu\textsuperscript{1,*}, Chenghao Liu\textsuperscript{2}, Zhuo Li\textsuperscript{3},  Jianling Sun\textsuperscript{1}\\
  \textsuperscript{1}Zhejiang University \\
  \texttt{\{linzhisheng, 11821003, sunjl\}@zju.edu.cn}\\
  \textsuperscript{2}Salesforce Research Asia \\ 
  \texttt{chenghao.liu@salesforce.com}\\
  \textsuperscript{3}State Street Technology (Zhejiang) Ltd. \\
  \texttt{lizhuo@zju.edu.cn}
}

\begin{document}
\maketitle

{
\renewcommand{\thefootnote}{{}}
\footnotetext{\textsuperscript{*}Corresponding author.}
}

\begin{abstract}
Parameter-efficient fine-tuning (PEFT) has emerged as an effective method for adapting pre-trained language models 
to various tasks efficiently. Recently, there has been a growing interest in transferring knowledge from one or multiple tasks to the downstream target task to achieve performance improvements. However,  current approaches typically either train adapters on individual tasks or distill shared knowledge from source tasks, failing to fully exploit task-specific knowledge and the correlation between source and target tasks. To overcome these limitations, we propose \name, a novel parameter-efficient fine-tuning framework based on multi-task transfer learning. \name extends the mixture-of-experts (MoE) framework to capture the transferable knowledge as a weighted combination of adapters trained on source tasks. These weights are determined by a gated unit, measuring the correlation between the target and each source task using task description prompt vectors. To fully exploit the task-specific knowledge, we also propose the Task Sparsity Loss to improve the sparsity of the gated unit. We conduct experiments on a broad range of tasks over 17 datasets. The experimental results demonstrate our \name yields stable improvements over full fine-tuning, and state-of-the-art PEFT and knowledge transferring methods on various tasks. The results highlight the effectiveness of our method which is capable of sufficiently exploiting the knowledge and correlation features across multiple tasks. Our code is available at \url{https://github.com/JachinLin2022/PEMT}
\end{abstract}
\section{Introduction}
Fine-tuning pre-trained models (PLMs) has become an effective way to migrate model capabilities to downstream tasks \cite{devlin2018bert, liu2019roberta, raffel2020exploring}. However, training and storing a full copy of the model parameters for each task becomes expensive as the scale of PLM increases. To mitigate this problem, parameter-efficient fine-tuning methods \cite{houlsby2019parameter,schick2021s,pfeiffer2020mad,lester2021power, liu2023gpt} have been proposed to reduce the number of trainable parameters. Despite their efficiency gains, these methods often sacrifice performance compared to full fine-tuning  \cite{gao-etal-2021-making,hu2021lora,li2021prefix}. 

\begin{figure}
  \centering
  \includegraphics[scale=0.5]{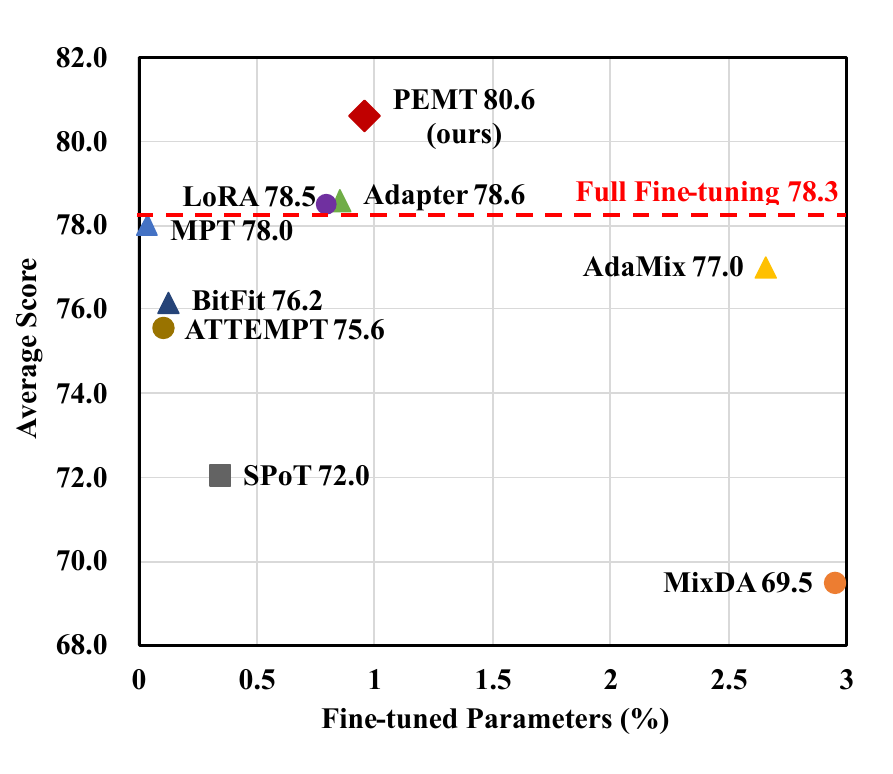}
  \caption{Performance of different parameter-efficient fine-tuning approaches. All results are based on T5-base for a fair comparison. 
  The proposed \name achieves significant improvements over all compared methods while fine-tuning only a small number of parameters.}
  \label{avgscore}
\end{figure}

Recent work has proposed to distill knowledge from one or multiple source tasks and adapt it to various downstream target tasks to achieve further improvements \cite{vu2021spot,asai2022attentional,wang2022multitask}. Despite significant success, there remains a substantial performance gap between these methods and full fine-tuning. The limitations of existing methods can be categorized as follows: (1) Most existing methods primarily focus on utilizing  shared knowledge across all source tasks, neglecting task-specific knowledge during adaptation to downstream tasks. (2) Task-specific representations of source and target tasks are typically trained independently, leading to insufficient exploitation of the correlation between them. As a result, the performance of multi-task transfer often lags behind that of distilling knowledge from a single source task. (3) The formulation of source and target tasks may be inconsistent, hindering cross-task adaptation. (4) The knowledge from source tasks is typically used as an initialization, but during fine-tuning, this knowledge may become intertwined with downstream tasks and gradually forgotten.

To mitigate these challenges, we propose \name, a \underline{p}arameter \underline{e}fficient fine-tuning framework based on \underline{m}ulti-task \underline{t}ransfer learning. \name comprises two training stages for source task learning and target task adaptation, respectively. (1) In Stage 1, we follow adapter-based tuning (\textit{e.g.}, Adapter \cite{houlsby2019parameter} and LoRA \cite{hu2021lora}) to train task-specific adapters on multiple source tasks. We incorporate a sequence of task-specific prompt vectors to distinguish different source tasks and utilize task descriptions to initialize each task prompt effectively. (2) In Stage 2, we train the adapter for the downstream task while incorporating knowledge from source tasks into the model. To enable multi-task transfer learning and prevent knowledge forgetting, we freeze the source adapters and integrate them using a mixture-of-experts architecture (MoE). Instead of relying on a single source task, the knowledge of source tasks is incorporated as a soft combination of all adapters trained during Stage 1. We employ an MoE gated unit to measure the correlation between the target task and each source task, leveraging the task-specific prompt vectors. To ensure the effective utilization of the specific knowledge from source tasks, we introduce the Task Sparsity Loss, encouraging the MoE gate to prioritize the most relevant source expert.

We conduct experiments on 17 NLP datasets involving multiple tasks and domains to evaluate the effectiveness of our approach. On all benchmarks, \name achieves an overall improvement of more than 2 points over full fine-tuning and all the compared PEFT methods as shown in Figure \ref{avgscore}. Under the few-shot setting, \name also proves a significant improvement of 10 points over the compared transfer learning models. Further analysis on the weights of different task experts demonstrates that the model tends to incorporate knowledge from the most relevant source task expert, which explains the efficiency and adaptability of our method.

Overall, this work makes the following contributions:
\begin{itemize}[leftmargin=*]
	\item We propose \name, a two-stage parameter-efficient fine-tuning method facilitating multi-task transfer learning. \name captures the transferable knowledge through a combination of adapters trained on source tasks, effectively leveraging task-specific knowledge.
	\item We propose a task-correlation-based gated unit to determine the weight of each source adapter by measuring the correlation between source and target downstream tasks. To capture interdependency across tasks, we introduce a sequence of task-specific prompt vectors to describe each task.
	\item Experimental results indicate \name consistently outperforms full fine-tuning and state-of-the-art PEFT methods across a broad range of tasks, which demonstrates the robustness and adaptability of our method. \name is proven to be also effective for few-shot learning using 4-32 labels. 
	\item We also conduct extensive experiments to analyze how the performance changes under various settings, which provides a clear interpretation for the effectiveness of the proposed method.
\end{itemize}

\section{Related Work}
\paragraph{Parameter-Efficient Fine-tuning.}
Parameter-efficient fine-tuning freezes the original PLM and introduces a small number of additional parameters for fine-tuning. Existing works can be categorized into two classes, adapter-based tuning and prompt-based tuning. Adapter-based methods \cite{houlsby2019parameter,pfeiffer2020mad} incorporate a trainable bottleneck module to each transformer layer. Prompt-based tuning \cite{lester2021power,schick2021exploiting,gao2021making} prepends continuous or discrete prompt vectors to the input.
Recently, some methods are proposed \cite{pfeiffer2021adapterfusion,vu2021spot,wang2022adamix,gururangan2022demix,diao2023mixture,he2022hyperprompt,asai2022attentional,wang2022multitask,zhao2023prototype} to transfer knowledge of trained adapters to downstream tasks.

\paragraph{Multi-Task Transfer Learning.}
Transferring knowledge from tasks has been proven to be an effective approach \cite{10.1145/3219819.3220007, aghajanyan2021muppet, zhong2021adapting, clark2019bam, singh2022frustratingly,gupta2022sparsely,frohmann2023scalearn}. Many studies \cite{sanh2022multitask, wei2021finetuned,wang2022benchmarking, liu2022few} show the zero-shot or few-shot transferring capabilities of language models through massive multi-task training over a broad range of tasks. However, the corresponding overhead could be enormous. To overcome the issue, some more recent works \cite{vu2021spot,asai2022attentional,wang2022multitask, diao2023mixture} propose to transfer the knowledge shared by various tasks using parameter-efficient fine-tuning.


Among the related works, AdaMix \cite{wang2022adamix} and MPT \cite{wang2022multitask} are the most relevant methods. Compared to our \name, AdaMix trains the representation of source and target tasks independently and fails to sufficiently leverage the interdependency across tasks. MPT learns a single prompt by distilling the shared knowledge while ignoring the rich task-specific information.
\section{Approach}
\paragraph{Task.} Given a set of $\mathcal{K}$ source tasks $\bm{ \mathcal{S}}=\{\mathcal{S}_1,\mathcal{S}_2,$ $\cdots,\mathcal{S}_\mathcal{K}\}$ and a set of $\mathcal{M}$ target tasks $\bm{\mathcal{T}}=\{\mathcal{T}_1,\mathcal{T}_2,\cdots,\mathcal{T}_\mathcal{M}\}$, our goal is to capture the knowledge of $\bm{\mathcal{S}}$ and adapt it to any target task $\mathcal{T}_{m} \in \bm{\mathcal{T}}$.

\begin{figure}
  \centering
  \includegraphics[width=1\linewidth]{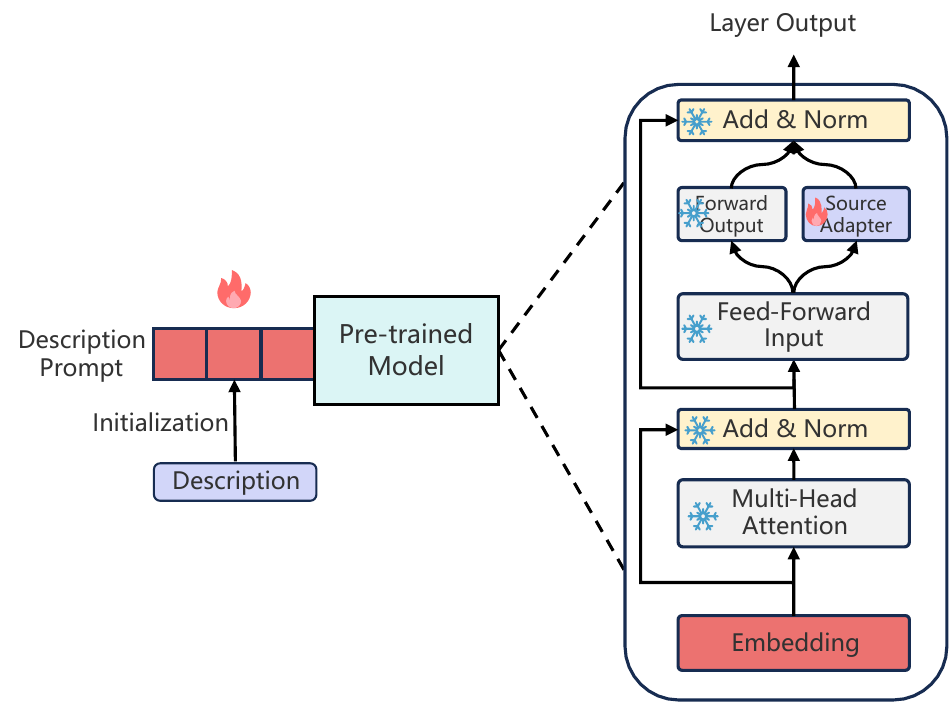}
  \caption{The training process of Stage 1. The task-specific adapters and task representation prompts are trained on multiple source tasks.}
  \label{overall-fig1}
\end{figure}

\begin{figure*}
  \centering
  \includegraphics[width=1\linewidth]{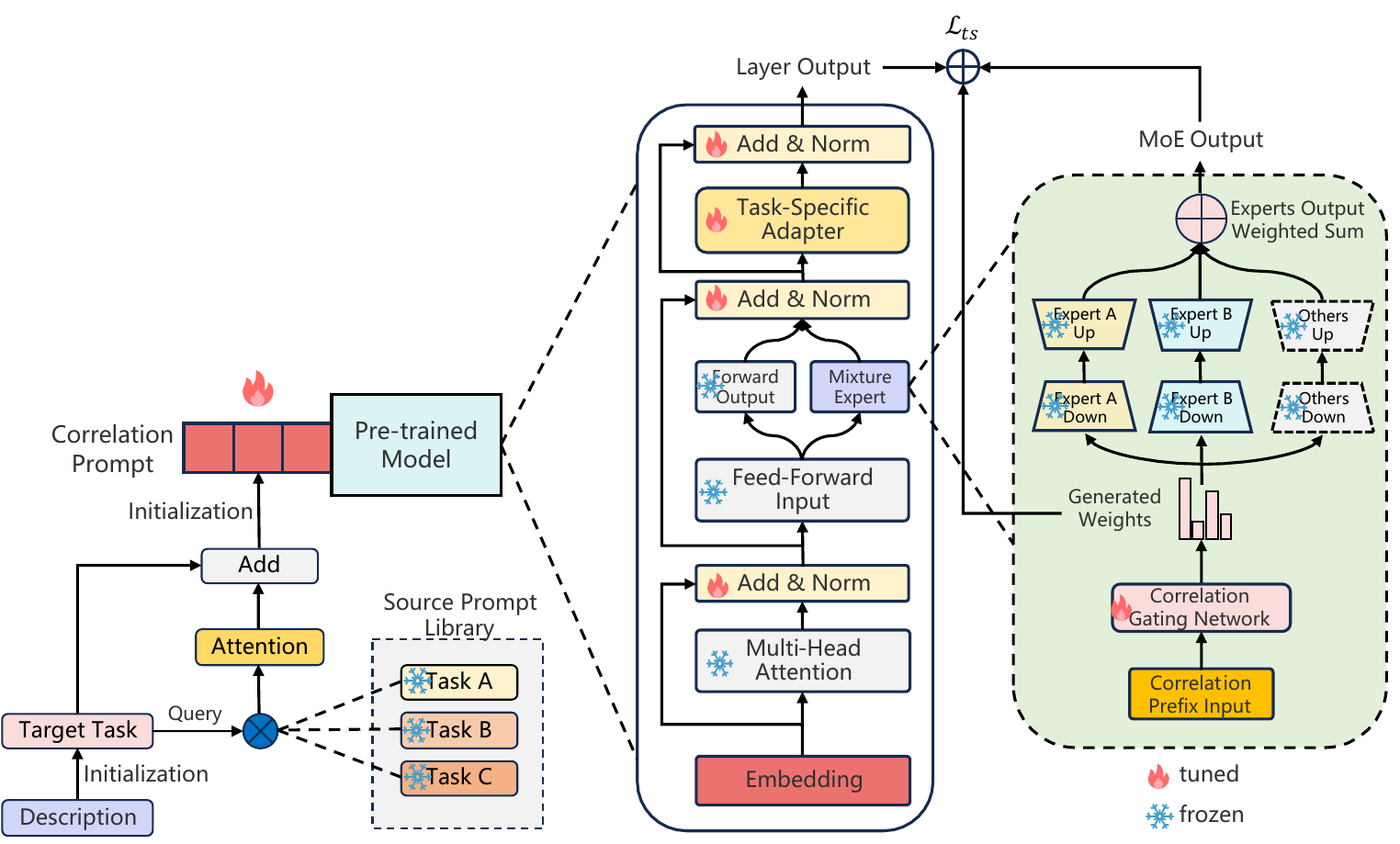}
  \caption{The training process of Stage 2. A MoE module is employed to distill knowledge from source tasks. The source task adapters are used as the experts and combined with a MoE gate which measures the correlation between the target task and each source task. The specific adapter for the target task is injected \textit{after} the MoE module. The task sparsity loss $\mathcal{L}_{ts}$ is incorporated to improve the sparsity of the MoE gate. The task prompt for the target task is a task-correlation-based combination of the trainable prompt vectors and the frozen prompts of the source tasks. }
  \label{overall-fig2}
\end{figure*}

\paragraph{Overview.} To sufficiently exploit the task-specific knowledge of each source task, we divide the training process of \name into two stages, which are illustrated in Figure \ref{overall-fig1} and Figure \ref{overall-fig2} respectively. In the first stage, we follow the vanilla adapter-based (\textit{e.g.}, LoRA \cite{hu2021lora} or Adapter \cite{houlsby2019parameter}) to train the source task adapters. For each source task, we freeze the original PLM parameters and inject a task-specific adapter to the feed-forward layer (FFN) for each Transformer layer. Besides, to learn a better representation of each task, we incorporate a task-specific description prompt which is used to measure the correlation between tasks. In Stage 2, we distill the knowledge of source tasks as a weighted combination of the source adapters. The Mixture-of-experts architecture (MoE) is exploited to integrate the frozen source adapters and the MoE gate measures the weight of each source expert using the task prompts learned in Stage 1. The task prompt for the target task is a correlation-based combination of the trainable prompt vectors and the frozen prompts of the source tasks. To adapt to a downstream task, another task adapter is injected \textit{after} the MoE module as shown in Figure \ref{overall-fig2}.


\subsection{Source Training}
The goal of Stage 1 is to capture the task-specific knowledge of each source task. To this end, we fine-tune the PLM on multiple source tasks using adapter-based PEFT methods.

\paragraph{Source Task Adapter.} 
As shown in Figure \ref{overall-fig1}, the task adapter is injected in each transformer layer, which works parallel to the FFN layer to learn the task-specific knowledge. This design is inspired by recent studies \cite{geva2021transformer, de2021editing, meng2022locating} that FFN captures the major knowledge of the training data.
To be specific, a transformer FFN consists of two stacked layers, an up projection layer and a down projection layer. We integrate an adapter module to the FFN using either a parallel Adapter \cite{he2021towards} or a LoRA, which works parallel to the up projection layer. The task adapter is implemented as two stacked low-rank matrices for reducing overheads.

\paragraph{Task Description Prompts.}
\label{sec:task_description_prompts}
We introduce a task description prompt for each source task. The prompt describes the task formulation and is utilized to measure the correlation between tasks. Existing methods \cite{vu2021spot,asai2022attentional,wang2022multitask} train the representations of various tasks from scratch independently, which brings a gap between tasks. To address this issue, we propose a simple but effective method to use handcrafted task descriptions as the initialization for the prompt vectors. Concretely, given a source task $\mathcal{S}_{k} \in \bm{\mathcal{S}}$, we prepend a trainable prompt matrix $\mathbf{P}_{k}\in\mathbb{R}^{N_k \times d}$ to the input tokens of PLM, where $d$ is the embedding dimension and $N_k$ denotes the length of the task description (\textit{i.e.} prompt length). The task description is a sentence consisting of a task definition and input-output format based on the distinctive features of various tasks. It should be noted that the description length for different tasks could be different. The details for the template of the task descriptions are provided in Appendix \ref{appendix:Description}.

\paragraph{Training on Source Tasks.}
Both the task adapters and task description prompts are trained following the typical PEFT procedure. There is no particular requirement for the source tasks. To bridge the gap between different tasks, we uniformly formulate all source tasks as text-to-text generation. We follow the format as proposed by \citet{raffel2020exploring}.

\subsection{Target Adaptation}
In the second stage, \name is guided by the correlation between tasks to utilize the distilled knowledge of all source tasks for adaptation to the downstream target task.

\paragraph{Mixture of Source Task Adapters.}
We employ a Mixture-of-Experts (MoE) module to combine the source adapters as the transferable knowledge. Instead of only focusing on the shared knowledge, we maintain the task-specific information of each source task during adapting to the downstream task.  As illustrated in Figure \ref{overall-fig2}, the task adapters trained in Stage 1 are exploited as the experts in the MoE module. Instead of fine-tuning the source task adapters, we freeze the parameters of the experts to avoid the catastrophic forgetting problem. Formally, the output of the MoE module in the $l$-th layer is calculated as:
\begin{equation}
	\mathbf{H}_{e}^{l} = \sum_{k=1}^{\mathcal{K}}w_{k}^{l}\cdot\mathbf{E}_{k}^{l},
\end{equation}
where $\mathbf{E}_{k}^{l}$ is the task adapter in the $l$-th Transformer layer trained on the $k$-th source task $\mathcal{S}_k$, and $\mathcal{K}$ is the total number of source tasks. $w_{k}^{l}$ denotes the weight of $\mathbf{E}_{k}^{l}$, which is obtained by the MoE gate, calculated as:
\begin{equation}
	w_k^l = {\rm softmax}\left(\mathbf{W}_g^l \cdot {\rm avg}\left(\mathbf{H}\right)\right)_k,
	\label{eqn:task_corr}
\end{equation}
where $\mathbf{W}_g^l \in \mathbb{R}^{d \times \mathcal{K}}$ is a trainable matrix, and $\rm avg$ is an average pooling layer. $\mathbf{H}$ is the prompt matrix for the current target task, which captures the correlation between tasks.

\paragraph{Correlation-Guided Task Prompt.}
As aforementioned, existing methods train source and target representations independently, which leaves an obstacle to acquire knowledge interdependency across tasks. To exploit the correlation between tasks sufficiently, we propose to incorporate the prompts trained on source tasks into the target adaptation process based on attention mechanism. Following \cite{vaswani2017attention}, the attention function ${\rm attn}(Q, K, V)$ takes three inputs, query, key, and value respectively. Here, we utilize the target prompt as a query and the source prompts as key and value. Formally, let $\mathbf{Q}=\left(\mathbf{q}_1, \cdots, \mathbf{q}_T\right)$ denotes the trainable prompt matrix of the target task. $\mathbf{Q} \in \mathbb{R}^{T \times d}$ is initialized with a task description of $T$ tokens following the same way as in Stage 1 and $\mathbf{q}_t$ denotes the $t$-th prompt vector. Given the prompt of the $k$-th source task $\mathbf{P}_k$, the correlation feature between source task $\mathcal{S}_k$ and the target task is obtained as:
\begin{equation}
	\mathbf{C}_k = {\rm attn}(\mathbf{Q}, \mathbf{P}_k, \mathbf{P}_k) \in \mathbb{R}^{T \times d}.
\end{equation}
Once the correlation feature for each source task is obtained, we simply add all the correlation information to the original prompt of the target task. Concretely, the final prompt matrix $\mathbf{H} \in \mathbb{R}^{T \times d} $ for the target task is calculated by: 
\begin{equation}
	\mathbf{H} = \mathbf{Q} + \sum_{k=1}^{\mathcal{K}}\mathbf{C}_k.
\end{equation}
This design is inspired by the additive compositionality of word embedding \cite{mikolov2013efficient}, which is proven to be simple but effective according to the experimental results. The prompt for the target task captures the representation information and interdependency across source and target tasks, and is exploited to measure the weight of each source task adapter (Eq \ref{eqn:task_corr}). It should be noted that all Transformer layers of \name share the same $\mathbf{H}$ for the sake of efficiency.

\paragraph{Target Task Adapter.}
To adapt to the downstream task, we incorporate another task-specific adapter into each Transformer layer. The target task adapter, which is inserted \textit{after} the MoE module, is exploited to mine the knowledge which is not covered by the experts trained on source tasks. The combination of source and target adapters facilitates the model to take advantage of both the rich knowledge learned from each source task and the task-specific knowledge of the target task.

\paragraph{Fine-Tuning on the Target Tasks}
To sufficiently utilize the knowledge of the source tasks, we propose the Task Sparsity Loss (TSL) to improve the sparsity of the MoE module. The intuition is to ensure the MoE gate assigns a higher priority to the top-1 source task expert by measuring the similarity between specific expert output and the final layer output. Formally, the TSL is defined as:
\begin{equation}
	\mathcal{L}_{ts} = -\frac{1}{L \mathcal{K}}\sum_{l=1}^{L}\sum_{k=1}^{\mathcal{K}}w_k^l\cdot {\rm sim}(\mathbf{H}_o^l, \mathbf{E}_{k}^l),
\end{equation}
where $\mathbf{H}_o^l$ denotes the final hidden state of the $l$-th Transformer layer, $L$ is the total number of layers, and ${\rm sim}$ is a similarity score function and we choose cosine similarity in this paper.

Similar to the training process on source tasks, we formulate the target tasks as a text-to-text generation problem.  The training objective is to minimize the negative log-likelihood of output $\mathbf{y}$ conditioned on the input text $\mathbf{x}$ and the task prompt $\mathbf{H}$. Finally, the fine-tuning loss on the target task is defined as:
\begin{equation}
	\mathcal{L} = -\sum_{j}P(y_j|\mathbf{y}_{<j};\mathbf{x}, \mathbf{H}) + \alpha\mathcal{L}_{ts},
\end{equation}
where $\alpha$ is a hyperparameter to balance the losses.

\begin{table*}
	\centering
	\resizebox{\linewidth}{!}{
		\begin{tabular}{c|cccccccccc}
			\Xhline{1px}
			\multirow{2}{*}{\textbf{Method}} & \multicolumn{10}{c}{\textbf{GLUE \& SuperGLUE}} \\
			\cline{2-11}
			& \textbf{STS-B} & \textbf{MRPC} & \textbf{RTE} & \textbf{CoLA}  & \textbf{Multi}  & \textbf{BoolQ}  & \textbf{WiC} & \textbf{WSC} & \textbf{CB} & \textbf{Avg.}\\
			\Xhline{1px}
			FT & 89.7 & \textbf{89.1} & 71.9 & 61.8 & 72.8 & 81.1 & \textbf{70.2} & 59.6 & 85.7 & 75.8 \\
			PT & 89.5 & 68.1 & 54.7 & 10.6 & 58.7 & 61.7 & 48.9 & 51.9 & 67.9 & 56.9 \\
			BitFit & 90.9 & 86.8 & 67.6 & 58.2 & 74.5 & 79.6 & 70.0 & 59.6 & 78.6 & 74.0 \\
			Adapter & 90.7 & 85.3 & 71.9 & 64.0 & 75.9 & 82.5 & 67.1 & \textbf{67.3} & 85.7 & 76.7 \\
			LoRA & \textbf{91.1} & 86.8 & 74.1 & 61.5 & 75.2 & 81.8 & 69.2 & 65.4 & 85.7 & 76.7 \\
			SPoT & 90.0 & 79.7 & 69.8 & 57.1 & 74.0 & 77.2 & 48.9 & 51.9 & 67.9 & 68.5 \\
			ATTEMPT & 89.7 & 85.7 & 73.4 & 57.4 & 74.4 & 77.1 & 66.8 & 53.8 & 78.6 & 73.0 \\
			MPT & 90.4 & \textbf{89.1} & 79.4 & 62.4 & 74.8 & 79.6 & 69.0 & \textbf{67.3} & 79.8 & 76.9 \\
			MixDA & 90.8 & 88.2 & 66.9 & 60.8& 59.2 & 61.7 & 48.9 & 50.0 & 78.6 & 67.2 \\
			Adamix & 91.0 & 88.2 & 70.5 & 58.7 & 72.9 & 80.2 & 63.6 & 51.9 & 85.7 & 73.6 \\
			\hline
			\textbf{\name} & \textbf{91.1}\textsubscript{0.22} & 88.7\textsubscript{0.40} & \textbf{83.0}\textsubscript{1.36} & \textbf{67.0}\textsubscript{2.12} & \textbf{75.5}\textsubscript{0.36} & \textbf{82.6}\textsubscript{0.38} & 68.7\textsubscript{0.89} & \textbf{67.3}\textsubscript{0.0} & \textbf{94.1}\textsubscript{1.68} & \textbf{79.8}\textsubscript{0.17} \\
			\Xhline{1px}
		\end{tabular}
	}
 
	\caption{\label{glue}
		Results on GLUE and SuperGLUE. The metrics are Pearson correlation for STS-B, F1 for MultiRC (Multi), and accuracy for other tasks as evaluation metrics. Our results are averaged over three runs, and subscripts denote standard deviation.
	}
\end{table*}
\section{Experiment}
We conduct experiments on a comprehensive range of NLP datasets to demonstrate the effectiveness of \name. The performance of different methods is compared under both full-dataset and few-shot settings.

\subsection{Datasets and Tasks}
As in \citet{wang2022multitask}, we use 6 high-resource datasets as the source tasks: MNLI \cite{williams-etal-2018-broad}, QNLI \cite{demszky2018transforming}, QQP \cite{wang-etal-2018-glue}, SST-2 \cite{socher-etal-2013-recursive}, SQuAD \cite{rajpurkar-etal-2016-squad}, and ReCoRD \cite{zhang2018record}. We use other datasets from four benchmarks as target tasks: MultiRC \cite{khashabi-etal-2018-looking}, BoolQ \cite{clark-etal-2019-boolq}, WiC \cite{pilehvar2019wic}, WSC \cite{levesque2012winograd} and CB \cite{de2019commitmentbank} from SuperGLUE \cite{wang2019superglue}; RTE \cite{giampiccolo2007third}, CoLA \cite{warstadt-etal-2019-neural}, STS-B \cite{cer2017semeval}, MRPC \cite{dolan-brockett-2005-automatically} from GLUE \cite{wang-etal-2018-glue}; Natural Questions (NQ) \cite{kwiatkowski-etal-2019-natural}, HotpotQA (HP) \cite{yang-etal-2018-hotpotqa}, NewsQA (News) \cite{trischler-etal-2017-newsqa}, and SearchQA (SQA) \cite{dunn2017searchqa} from MRQA \cite{fisch2019mrqa}; WinoGrande \cite{sakaguchi2021winogrande}, Yelp-2 \cite{zhang2015character}, SciTail \cite{khot2018scitail}, and PAWS-Wiki \cite{zhang-etal-2019-paws} from the \textit{Others} benchmark as in \cite{asai2022attentional}.
\paragraph{Compared Methods}
We compare \name with the state-of-the-art fine-tuning methods: (1) Full fine-tuning (FT), which fine-tunes all parameters of the pre-trained model. (2) Prompt-based tuning, including vanilla prompt tuning (PT) \cite{lester2021power}, SPoT \cite{vu2021spot}, ATTEMPT \cite{asai2022attentional} and MPT \cite{wang2022multitask}. (3) Adapter-based tuning, including vanilla adapter \cite{houlsby2019parameter}, AdaMix \cite{wang2022adamix} and MixDA \cite{diao2023mixture}. (4) Other parameter-efficient tuning methods, including LoRA \cite{hu2021lora} and BitFit \cite{zaken2022bitfit}.

  \begin{table*}
	\centering
	\scalebox{0.86}{
		\begin{tabular}{c|ccccc|ccccc}
			\Xhline{1px}
			\multirow{2}{*}{\textbf{Method}} & \multicolumn{5}{c|}{\textbf{MRQA}} & \multicolumn{5}{c}{\textbf{Others}} \\
			\cline{2-11}
			& \textbf{NQ} & \textbf{HP} & \textbf{SQA} & \textbf{News}  & \textbf{Avg.}  & \textbf{WG}  & \textbf{Yelp} & \textbf{SciTail} & \textbf{PAWS} & \textbf{Avg.}\\
			\Xhline{1px}
			FT & \textbf{75.1} & 77.5 & 81.1 & 65.2 & 74.7 & 61.9 & 96.7 & 95.8 & 94.1 & 87.1 \\
			PT & 67.9 & 72.9 & 75.7 & 61.1 & 69.4 & 49.6 & 95.1 & 87.9 & 55.8 & 72.1 \\
			BitFit & 70.7 & 75.5 & 77.7 & 64.1 & 72.0 & 57.2 & 94.7 & 94.7 & 92.0 & 84.7 \\
			Adapter & 74.2 & 77.6 & 81.4 & 65.6 & 74.7 & 59.2 & 96.9 & 94.5 & 94.3 & 86.2 \\
			LoRA & 73.9 & 77.1 & 80.1 & 64.9 & 74.0 & 60.2 & 96.4 & 94.5 & 94.2 & 86.3 \\
			SPoT & 68.2 & 74.8 & 75.3 & 58.2 & 69.1 & 50.4 & 95.4 & 91.2 & 91.1 & 82.0 \\
			ATTEMPT & 70.4 & 75.2 & 77.3 & 62.8 & 71.4 & 57.6 & 96.7 & 93.1 & 92.1 & 84.9 \\
			MPT & 72.0 & 75.8 & 77.2 & 63.7 & 72.2 & 56.5 & 96.4 & 95.5 & 93.5 & 85.5 \\
			MixDA & 71.2 & 76.1 & 78.3 & 63.9 & 72.4 & 55.2 & 95.7 & 50.8 & 82.7 & 71.1 \\
			Adamix & 73.2 & 77.5 & 80.4 & 65.2 & 74.1 & 59.8 & 96.6 & 96.0 & 94.0 & 86.6 \\
			\hline
			\textbf{\name} & $\textbf{75.1}\textsubscript{0.04}$ & $\textbf{78.3}\textsubscript{0.10}$ & $\textbf{81.8}\textsubscript{0.09}$ & $\textbf{65.9}\textsubscript{0.12}$ & $\textbf{75.3}\textsubscript{0.02}$ & $\textbf{62.3}\textsubscript{0.08}$ & $\textbf{97.0}\textsubscript{0.06}$ & $\textbf{96.9}\textsubscript{0.69}$ & $\textbf{94.3}\textsubscript{0.08}$ & $\textbf{87.6}\textsubscript{0.18}$ \\
			
			\Xhline{1px}
		\end{tabular}
	}
	\caption{\label{mrqa}
		Results on MRQA and the \textit{Others} benchmark. 
		Our results are averaged over three runs and subscripts indicate standard deviation. 
	}
\end{table*}

\begin{table*}
	\centering
	\scalebox{0.86}{
		\begin{tabular}{c|c|cccccccccc}
			\Xhline{1px}
			\multirow{2}{*}{\textbf{$k$-shot}} & \multirow{2}{*}{\textbf{Method}} & \multicolumn{10}{c}{\textbf{GLUE \& SuperGLUE}} \\
			\cline{3-12}
			& & \textbf{STS-B} & \textbf{MRPC} & \textbf{RTE} & \textbf{CoLA}  & \textbf{Multi}  & \textbf{BoolQ}  & \textbf{WiC} & \textbf{WSC} & \textbf{CB} & \textbf{Avg.}\\
			\hline    
			\multirow{3}{*}{4}&PT & 88.8 & 68.1 & 56.3 & 27.4 & 61.8 & 61.6 & 51.2 & 60.4 & 53.5 & 58.8 \\
			&MPT & 89.1 & 68.1 & 62.6 & 34.8 & 62.2 & 62.2 & 52.9 & \textbf{67.3} & 73.6 & 63.6 \\
			&\textbf{\name} & \textbf{89.2} & \textbf{78.4} & \textbf{64.0} & \textbf{44.7} & \textbf{72.0} & \textbf{71.0} & \textbf{62.1} & 44.2 & \textbf{78.6} & \textbf{67.1} \\
			\hline
			\multirow{3}{*}{16}&PT & 87.8 & 68.1 & 54.7 & 28.5 & 60.3 & 61.9 & 48.9 & 44.2 & 63.5 & 57.5 \\
			&MPT & 89.1 & 70.1 & 64.8 & 32.1 & 64.5 & 63.3 & 49.8 & \textbf{67.3} & 78.6 & 64.4 \\
			&\textbf{\name} & \textbf{89.8} & \textbf{86.8} & \textbf{69.8} & \textbf{43.4} & \textbf{72.4} & \textbf{74.0} & \textbf{66.5} & 44.2 & \textbf{82.1} & \textbf{69.9} \\
			\hline
			\multirow{3}{*}{32}&PT & 87.5 & 68.1 & 54.7 & 23.2 & 59.2 & 61.7 & 52.6 & \textbf{67.3} & 67.8 & 60.2 \\
			&MPT & 89.7 & 74.5 & 59.7 & 30.8 & 63.3 & 68.9 & 53.9 & \textbf{67.3} & 82.1 & 65.6 \\
			&\textbf{\name} & \textbf{89.8} & \textbf{86.3} & \textbf{71.9} & \textbf{45.5} & \textbf{72.2} & \textbf{74.4} & \textbf{61.8} & 51.9 & \textbf{85.7} & \textbf{71.1} \\
			\Xhline{1px}
		\end{tabular}
	}
	\caption{
		Few-shot learning results on GLUE with 4, 16, and 32 training examples. 
	}
	\label{few-shot}
\end{table*}

\subsection{Implementation}
Following existing works, we use the publicly available pre-trained T5-Base model \cite{raffel2020exploring} with 220M parameters from HuggingFace\footnote{https://huggingface.co/} as the backbone. 

Following \cite{karimi2021compacter}, if a dataset does not have a publicly available test split with annotations, we use the full set of a subset of the developing partition or a subset of the for testing. \name is trained on 4 x NVIDIA A800 GPUs. The implementation details and hyper-parameters are listed in Appendix \ref{appendix:Implementation}.

We run all the experiments three times with different random seeds, and report the mean values and standard deviations. Under the few-shot setting, for each number of shots $k \in \{4, 16, 32\}$, we randomly collect $k$ samples from the downstream task data. The random seed is shared by all compared methods for a fair comparison.

\subsection{Results}
\paragraph{Full Data.} Experimental results in Table \ref{glue} and \ref{mrqa} show that \name significantly outperforms full fine-tuning and all other parameter-efficient tuning methods.
As observed from Table \ref{glue}, \name establishes the new state-of-the-art results for parameter-efficient fine-tuning on GLUE and SuperGLUE. According to the results, Adapter and MPT are the most competitive methods, while our method yields an improvement of 2.75\% and 2.91\%. Especially, On CB task, the improvement comes to 13.06\% and 7.16\%. On RTE task, \name outperforms all other methods with over 10 points, which illustrates the capability of knowledge transferring of our method. 

Table \ref{mrqa} shows the performance of different methods on MRQA and \textit{Others} benchmark. Compared with GLUE and SuperGLUE, the data sizes of these two datasets are larger, and the contexts of the samples are longer. Due to these complexities, the performance of previous PEFT methods is significantly inferior to full fine-tuning. From the results, \name successfully outperforms full fine-tuning on these datasets, suggesting the stability and robustness of \name across different data sizes and context lengths.

\paragraph{Few-shot.} Following prior works, we conduct few-shot experiments on GLUE and SuperGLUE benchmark to measure the generalization of \name to new tasks with only a few training examples available ($k \in \{4, 16, 32\}$).
Table \ref{few-shot} shows the results. With limited data resources, our method still yields a significant improvement, especially on some tasks such as WiC, MultiRC, CoLA, and MRPC.
Another interesting observation is that the improvement of \name over baselines becomes more pronounced as the number of training samples increases. This further underscores that the task-shared knowledge of MPT gradually fades during the training process of downstream tasks when more training data is provided. In contrast, \name freezes the source task adapters, which not only preserves shared knowledge to the greatest extent possible but also sufficiently exploits the associations and distinctions across various tasks.


 \begin{figure*}
	\centering
	\includegraphics[width=1\linewidth]{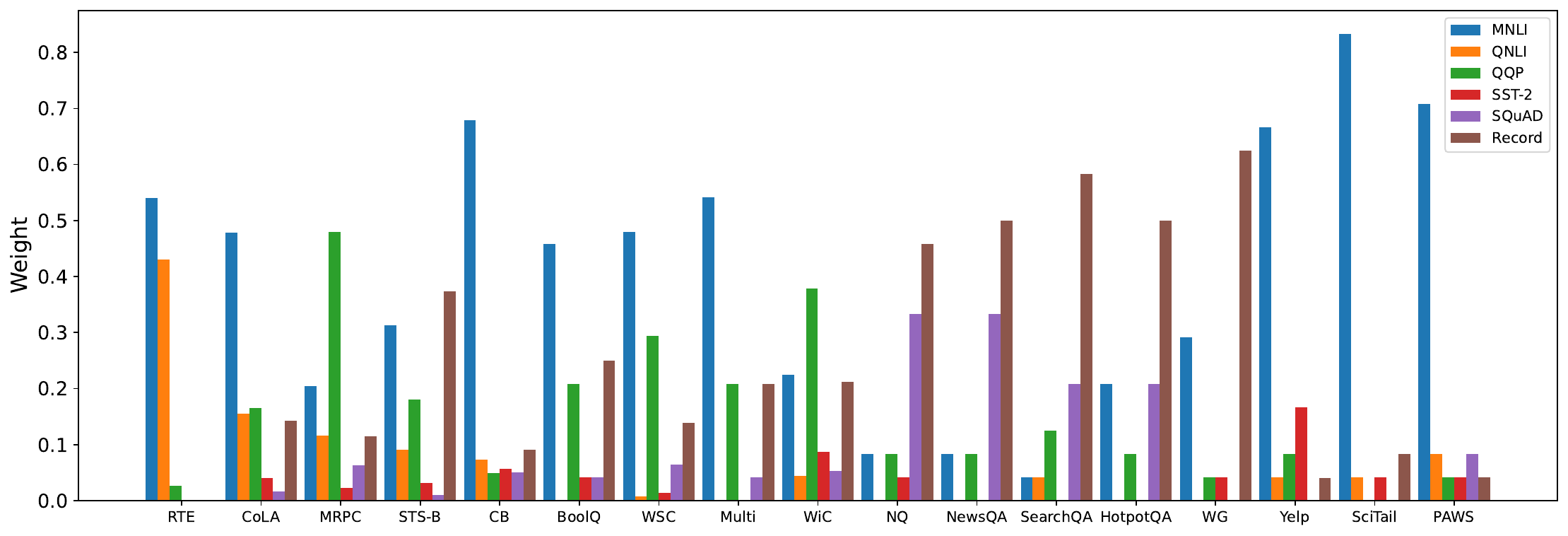}
	\caption{The source expert weight distribution in GLUE, SuperGLUE, MRQA and \textit{Others} benchmarks.}
	\label{histogram1}
\end{figure*}

\begin{table*}[h]
	\centering
	\scalebox{0.9}{
		\begin{tabular}{c|cccccccccc}
			\Xhline{1px}
			\textbf{Number of Source Task}& \textbf{STS-B} & \textbf{MRPC} & \textbf{RTE} & \textbf{CoLA}  & \textbf{Multi}  & \textbf{BoolQ}  & \textbf{WiC} & \textbf{WSC} & \textbf{CB} & \textbf{Avg.} \\
			\hline
			1 & 91.3 & 86.8 & 80.6 & 66.3 & 76.1 & 81.5 & 65.4 & 67.3 & 92.9 & 78.7 \\
			2 & 91.0 & 87.8 & 83.5 & 64.3 & 75.4 & 82.1 & 68.3 & 67.3 & 92.9 & 79.2 \\
			4 & 91.3 & 89.2 & 82.0 & 64.2 & 74.9 & 82.6 & 68.7 & 67.3 & 92.9 & 79.2 \\
			6 & 91.1 & 89.7 & 83.0 & 67.0 & 75.5 & 82.6 & 68.7 & 67.3 & 94.1 & 79.8 \\
			\Xhline{1px}
		\end{tabular}
	}
	\caption{Average scores on GLUE and SuperGLUE benchmark with different number of source tasks.}
	\label{tab:scores}
\end{table*}

\begin{table}[h]
    \centering
    \scalebox{0.9}{
        \begin{tabular}{ccccc}
        \Xhline{1px}
        \textbf{Method} & \textbf{Train(s)} & \textbf{Infer(s)} & \textbf{Mem(GB)} \\
        \Xhline{1px}
            FT & 493.17 & 0.78 & 8.57 \\
            LoRA & 391.60 & 0.76 & 5.28 \\
            Adapter & 377.00 & 0.79 & 5.74 \\
            MPT & 2163.30 & 5.53 & 50.57 \\
            ATTEMPT & 2177.41 & 5.58 & 50.86 \\
        \hline
        \multicolumn{5}{c}{\textbf{Number of Source Task for PEMT}} \\
        \hline
            1 & 364.35 & 0.73 & 5.56 \\
            2 & 433.96 & 0.85 & 8.78 \\
            4 & 478.60 & 0.95 & 10.77 \\
            6 & 551.39 & 1.11 & 12.64 \\
        \Xhline{1px}
        \end{tabular}
    }
    \caption{Comparison of efficiency on CoLA task.}
    \label{tab:cola_comparison}
\end{table}
\section{Analysis}
We conduct further analysis to investigate the effectiveness of different components of \name.
\paragraph{Weights of Source Adapters.}
In order to explore how the weights of source experts change on various target tasks, we collect the outputs of the MoE gate and visualize them through histograms as shown in Figure \ref{histogram1}. As observed, there are obvious tendencies and priorities in the weight distribution. For GLUE and SuperGLUE benchmarks, the knowledge of the MNLI plays a dominant role, with a weighting of more than 50\% of all tasks. The contributions of some individual tasks are close to 0 under the constraint of Task Sparsity Loss. Contrastively, the distribution of weights on MRQA is totally different, where the two tasks SQuAD and ReCorD account for about 80\% of the weights. The reason is that all the three datasets MRQA, SQuAD and ReCorD belong to the Q\&A category, which also indicates the correlation guided MoE module and the task sparsity loss effectively work as expected.



\paragraph{Number of Source Tasks}
To substantiate the scalability of \name, we investigate how the performance changes when different numbers of source tasks are used in Stage 1. As shown in Table \ref{tab:scores}, compared to MixDA, \name exhibits a gradual improvement as the number of source tasks increases, which is different from the results of existing methods \cite{diao2023mixture}. This observation suggests the capability of \name to sufficiently capture the commonalities and differences among various tasks, which demonstrates a certain degree of continual learning proficiency. We also conducted efficiency experiments on the CoLA task, with specific experimental details provided in Appendix \ref{appendix:Implementation}. As shown in Table \ref{tab:cola_comparison}, compared to MPT and ATTEMPT, PEMT used a more concise prompt, achieving advantages in training, inference efficiency, and memory usage. However, we also observed that as the number of source tasks increased, the computational efficiency of PEMT gradually declined.

\paragraph{Task Description Prompts.}
As introduced in Section \ref{sec:task_description_prompts}, we initialize the task prompt with a sentence of task description. To measure the effectiveness of this method in maintaining consistency in task representation, we replace it with a randomly initialized prompt and keep the prompt length the same. As shown in Table \ref{ablation} (Row 2), the averaged score on the two benchmarks decreases by 0.8\% without initialization with task descriptions.

\begin{table}
  \centering
  \scalebox{1}{
  \begin{tabular}{ccc}
    \Xhline{1px}
    \textbf{No.} & \textbf{Ablation} & \textbf{Avg. Score} \\
    \Xhline{1px}
    1 & \name with LoRA & \bf 79.8 \\
    \hline
    2 & w/o description & 79.0 \\
    3 & w/o correlation  & 78.3 \\
    4 & w/o correlation and MoE & 76.6 \\
    \hline
    5 & \name with Adapter & 79.0 \\
  \Xhline{1px}
  \end{tabular}
  }
  \caption{\label{ablation}
  Results of ablation studies on GLUE and SuperGLUE benchmark.
  }
\end{table}

\paragraph{Correlation Guided Task Prompt.}
We conduct experiments to evaluate the effectiveness of task correlation features in facilitating the model to select the optimal source expert. We remove the entire prompt module in both source task training and target adaptation while maintaining the MoE module. For the MoE gate, we use an average pooling on the hidden states of the previous FFN layer as input. The ablation study in Table \ref{ablation} (Row 3) shows that task correlation features produce a 1.5\% average performance improvement.

\paragraph{Mixture-of-Source-Adapters.}
We further investigate the effectiveness of the source adapters on target adaptation.
To this end, we remove both the target prompt and the source adapters, while only maintaining the task-specific adapter \textit{after} each FFN layer. This change degenerates the model to the simple variant of Adapter which inserts an adapter module into each multi-head attention and FFN layer. We evaluate the performance on target adaptation without training on source tasks. The results in Table \ref{ablation} (Row 4) show that, without the task prompt and source adapters, the performance drops sharply by 3.2\% on average.
\section{Conclusion}
In this paper, we propose \name, a new parameter-efficient fine-tuning framework that is capable of adapting the knowledge from multiple tasks to the downstream target tasks. \name is facilitated with the correlation features between tasks and sufficiently leverages the task-specific knowledge of source tasks with prompt tuning and the mixture-of-experts architecture. We also introduce novel methods to improve prompt initialization and model sparsity. Experiments are conducted on a comprehensive range of datasets involving multiple tasks and domains and the results demonstrate \name significantly outperforms existing SOTA methods.

\section*{Limitations}
The model's inference latency rises proportionally with the number of experts, prompting the necessity to identify a stable reparameterization for merging the weights of multiple experts or to explore a reliable pruning method. Additionally, the entire framework involves a two-stage transfer learning process. Although this two-stage architecture significantly enhances performance on downstream tasks, it incurs substantial training overhead and data costs, and also introduces potential risks of data leakage or model attacks.

\bibliography{custom}
\bibliographystyle{acl_natbib}
\clearpage
\appendix

\section{Task Description Details}
We designed task descriptions based on the distinctive features of various tasks. Take MNLI task as an example, we use a description “Given a premise sentence and a hypothesis sentence, predict whether the premise entails the hypothesis, contradicts the hypothesis, or neither” to initialize the continuous prompt vectors prepended to the input. The descriptions for all tasks are shown as Table \ref{tab:description}.
\label{appendix:Description}

\begin{table}[h]
  \centering
  \begin{tabular}{c|c}
      \Xhline{1px}
      \textbf{Param} & \textbf{Value} \\
      \Xhline{1px}
      Optimizer & AdamW \\
      Learning rate & 5e-4 \\
      Batch size & 128 \\
      Warmup steps & 500 \\
      Expert dimension & 64 \\
      Training epochs & 5 \\
      Learning rate schedule & linear decay \\
      \Xhline{1px}
  \end{tabular}
  \caption{Stage 1 training: experimental setup.}
  \label{tab:stage1_param}
\end{table}

\begin{table}[ht]
  \centering
  \begin{tabular}{c|c}
      \Xhline{1px}
      \textbf{Param} & \textbf{Value} \\
      \Xhline{1px}
      Optimizer & AdamW \\
      Learning rate & \{6e-4, 1e-3\} \\
      Batch size & \{64, 128\} \\
      Expert dimension & 64 \\
      Training epochs & 20 \\
      Seed & \{42, 1024, 4096\} \\
      MoE loss factor & 0.1 \\ 
      Learning rate schedule & linear decay \\
      \Xhline{1px}
  \end{tabular}
  \caption{Stage 2 training: experimental setup.}
  \label{tab:stage2_param}
\end{table}

\section{Implementation Details}
We use down projection dimension $r=64$ in both source training and target adaptation. For source training, we train \name on each source task for 5 epochs. For target adaptation, we train all of the baselines for 20 epochs on small datasets with less than 10k examples, 10 epochs on medium size data with more than 10k examples, and 5 epochs on MRQA datasets. We limit the maximum training data number of Yelp-2 to be 100k samples. We run inferences on the test data using the model with the best development performance. We set the maximum token length to be 512 for MRQA datasets, 348 for MultiRC and 256 for all of other datasets. We set the maximum length of the input to be 256, 256, 512, 256 for GLUE, SuperGLUE, MRQA 2019, and \textit{Others} task set, respectively. We set the maximum length of input to be 348 for MultiRC. The details for training parameters are shown in Table \ref{tab:stage1_param} and Table \ref{tab:stage2_param}. For the efficiency experiments in Table \ref{tab:cola_comparison}, we meticulously measured the time and memory consumption during both the training and inference stages. All the efficiency experiments were conducted on a single A800 GPU with a batch size of 128 and precision set to fp32, with a prompt length of 100 to ensure consistency with their papers for MPT and ATTEMPT. We utilized the Transformers library to calculate GPU memory usage.
\label{appendix:Implementation}

\begin{table*}
  \centering
  \scalebox{0.9}{
  \begin{tabular}{l|p{15cm}}
      \Xhline{1px}
      \textbf{Task} & \textbf{Description} \\
      \Xhline{1px}
      QNLI & Given a question and a context sentence, determine whether the context sentence contains the answer to the question. \\
      MNLI & Given a premise sentence and a hypothesis sentence,  predict whether the premise entails the hypothesis (entailment), contradicts the hypothesis (contradiction), or neither (neutral). \\
      QQP & Given a pair of sentences, determine if the two sentences are semantically equivalent or not. \\
      SST-2 & Given a sentence,  predict whether a given sentence expresses a positive or negative sentiment. \\
      ReCoRD & Given a passage and a cloze-style question about the article in which one entity is masked out,  predict the masked out entity from a list of possible entities in the provided passage. \\
      SQuAD & Given an article and a corresponding question about the article,  answer the question accurately based on the provided context in the articles. \\
      \Xhline{1px}
      CoLA & Given a sentence, judge the grammatical acceptability of the sentence. \\
      RTE & Given a premise sentence and a hypothesis sentence, determine whether the hypothesis can be inferred from the premise. \\
      MRPC & Given a pair of sentences, determine whether the two sentences are semantically equivalent or not. \\
      STS-B & Given a pair of sentences,  measure the degree of semantic similarity or relatedness between pairs of sentences. \\
      CB & Given a premise and a hypothesis, determine the type and strength of the commitment being expressed. \\
      WiC & Given a target word and a pair of sentences, determine if a given target word in a sentence has the same meaning in two different contexts. \\
      WSC & Given a set of sentences that contain an ambiguous pronoun, determine the referent of the ambiguous pronoun based on the context provided. \\
      BoolQ & Given a question and a paragraph, determine if a given question can be answered with a simple "true" or "false" based on a given passage of text. \\
      Multi & Given a passage of text and a set of related multiple-choice questions, where each question is accompanied by several answer choices,  select the correct answer choice for each question based on the information provided in the passage. \\
      MRQA & Given an article and a corresponding question about the article, answer the question accurately based on the provided context in the articles. \\
      SciTail & Given a premise and a hypothesis,  classify the relationship between the premise and the hypothesis as entail or neutral. \\
      Yelp & Given a Yelp sentence,  predict the sentiment polarity (positive or negative) of customer reviews from the Yelp dataset. \\
      WG & Given a sentence and two options,  choose the right option for a given sentence which requires commonsense reasoning. \\
      PAWS & Given a pair of sentence, where one sentence is a paraphrase of the other. Determine if the given sentence pair is a paraphrase or not. \\
      \Xhline{1px}
  \end{tabular}
  }
  \caption{Tasks descriptions for prompt Initialization}
  \label{tab:description}
\end{table*}

\end{document}